# Microcontroller Based Robotic Arm Development for Library Management System

Bodhisatwa Barma[1], Samrat Ghosh[2*],
Abhrodip Chaudhury[3] and Biswarup Ganguly[4]

Department of Electrical Engineering
Meghnad Saha Institute of Technology,
Kolkata, India

theartofdeception08@gmail.com[1], iamsamratghosh@gmail.com[2],
abhrodip.chaudhury2010@gmail.com[3], bganguly@msit.edu.in[4]

**Abstract.** With the advancement of robotics, automation in various industries and processes has become widespread. This project aims to introduce library automation system, which addresses the fulfillment of the objectives of automatic retrieval of queued books, arrangement of returned books on the racks as well as automated updating of the library database. The proposed system is based on the Arduino microcontroller and python programming Microcontroller based robotic arms are used to fetch books from or return books to the different shelves in the library. The library database is also updated after completion of an action. The uniqueness of the proposed system lies in the fact that it can be applied to any existing library and is capable of handling individual books rather than a bulk. The system aims to bring new dimensions to the concept of library automation.

## 1 Introduction

The consequences of an improperly arranged library have been felt by almost all. May it be misplacing returned books or inability to find books being searched for, anything that is left in disarray comes back to haunt, at a later time. This project aims to solve this problem in a modern, technologically advanced and innovative way.

A robotic arm is used is to accept books from readers and put them back in their respective positions on racks. A controller is used to keep records of the books returned, sort the book according to some set algorithm, thus determining the address on the rack where the book will be placed by the arm. A library member can access the database at set platforms and any book requested for at this platform will be retrieved by the robotic arm in a reverse process. The system can have a wide range of implementations, the primary ones being medium and large sized libraries, where manual management becomes difficult. The versatility in the proposed system lies in the fact that it can be added to existing libraries rather than designing a unique stand-alone library. Researches have already been done in the field of library automation [1-4]. One example of an existing system is the automated storage & retrieval library system in Macquarie University, Sydney, Australia. It is an Radio Frequency Identification and Data collection (RFID) [5-6] based system which maintains a digital database of the books and queued books are retrieved by bringing an entire bin (container in which books of same category are arranged) to the reader. Story *et. al.* [7] have proposed a kiosk system with a walk-in enclosure, interactive selection panel and multi-section inventory storage area, that dispenses the books and accepts returns. Ostwald *et. al.* [8] have developed a system comprising a two-dimensional array that contains media cartridge storage cells and media cartridge players. In [9] ,Ehrenberg *et. al.* have presented LibBot, a robot equipped with an RFID reader, that automates the otherwise manual shelf-reading process and finds misplaced books autonomously. Dhanalakshmi *et. al.* [10] have proposed a system which allows identification of large number of tagged objects like books, using radio waves. The system consists of a transaction module that handles issue and return of books, and updates the database; a monitoring module at the gates which monitors incoming and outgoing bags continuously; and a searching module that provides navigational guidance to the users within the library to physically retrieve the resource. The system allows fast transaction flow for the library and will prove immediate and long term benefits to library in traceability and security. The main deficiencies of the systems with our proposed methodologies are as follows; The systems deal with one task at a time, i.e. either retrieval of books or arrangement of returned books. In this paper a more advanced system is designed to handle both tasks simultaneously while updating the library database. The commercially implemented systems cannot deal with individual book, rather they deal with blocks or stacks of several books. In our proposed work, individual book is dealt at a time that can be accessed by a single person at a time which can become time consuming.

## 2 Design of the system

The major components required for this project are: microcontroller (Arduino Mega 2560), RF Module (434 MHz), RGB scanner (2.7-3.3 V, 235 µA), IR LEDs and Piezo sensors (IR: $V_F$ = 1.3 V, $I_F$ = 50 mA, $V_{BR}$ = 60 V at $I_R$ = 100 µA Piezo: Easy Electronics EEPiezo25mm), Motors (150 rpm), Motor Drivers (L293, L298), Raw Materials for Mechanical Arm Construction (PVC, Wood, Copper-clad boards), Transformers (240/24 V), Diodes (6A4 series), Capacitors (220 µF, 35 V), Regulator (LM 317), Potentiometer (Wire Wound, 3W, 10K), Resistors (100 Ω, 270 Ω, 2 W).

DC power is required for the system to work. The power is supplied using a linear, regulated DC power supply (designed for the project) which is composed of a rectifier circuit, filter circuit and a regulator circuit.

Arm is supplied power via the ceiling mounted rails themselves. The two uppermost motors are used for movement of the arm. The operation of the motor that drives the pulley is similar to that of a hoist drive. The lower chamber that is hoisted from the cable has to be reinforced with collapsible bars to prevent swaying of this chamber during putting books in racks or due to wind. The lower chamber contains two motors, one for to-and-fro motion of the grabbing mechanism such that it can put books inside shelves, and a second motor for the grabber mechanism. Piezo sensors are mounted on this grabber so as to signal the microcontroller for proper operation of the grabber motor. The schematic diagram of the robotic arm is shown in Fig.1.

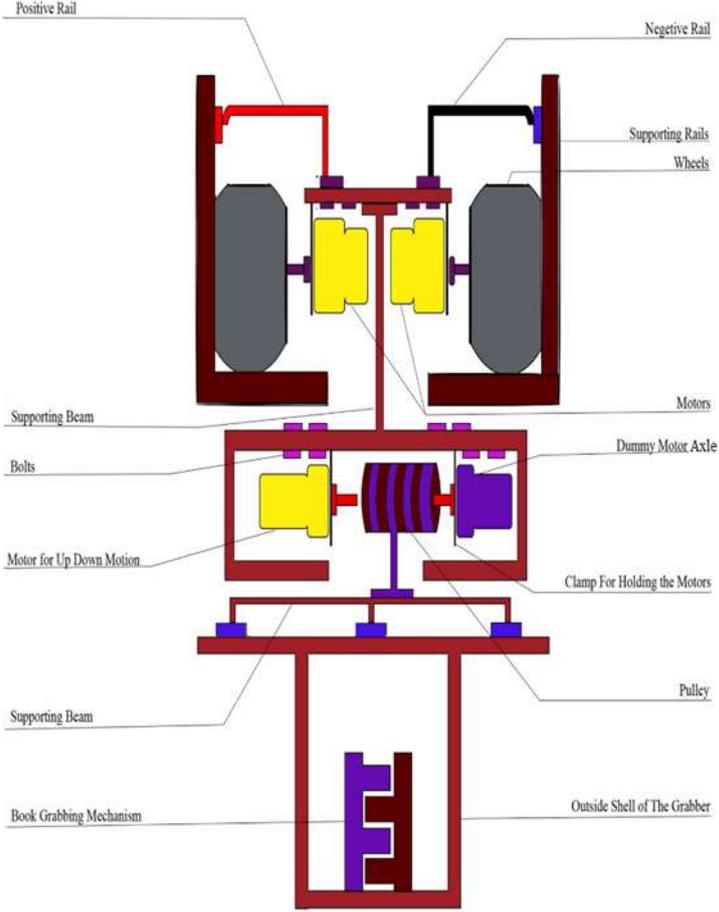

**Fig. 1.** Schematic diagram of robotic arm

## 3 Proposed Algorithm

The block diagram of the proposed work is shown in Fig. 2. The proposed system is based on a microcontroller that controls the action of some actuators based on a series of inputs. The principle by which books are accepted into the library and arranged on the shelves is discussed first. The process begins from a barcode scanner that scans every book that is returned to the library by the readers. Considering the case of one particular book, the input from the barcode scanner is sent to the microcontroller via an RF module. The microcontroller then processes this information, and sorts the book according to a set algorithm (based on perhaps alphabetical arrangement according to title, author, genre etc.) and determines a unique address for it on the shelves. The arm then takes the book to this address, the grabbing process being controlled by inputs from piezo sensors mounted on the grabber to account for varying widths. The arm can either be mounted on the floor directly, or on a set of rails hung from the ceiling. The ceiling mounted rail system is used here as the floor mounted system has some difficulty in case of tightly packed libraries or in case of high shelves. The direction of movement of the arm is changed using rotating turn-tables, which can either be operated by the arm itself, or be separately operated by a set of microcontrollers, the latter scheme is chosen for this project. This scheme is chosen to set up a controllable network of rails that can accommodate and efficiently manage the movement of multiple arms in a same library simultaneously (similar to a railway network). On the shelves, there are mechanical grabbers (either, spring operated tong like structures; or alternatively, simple slot arrangement) arranged at optimal distances which serve to hold the books in their place, so that they do not fall and consequently, obstruct the placement of any other book. Each of these 'positions' at which these grabbers are located are the addresses which the books are kept at. The optimal distance between successive grabbers is so adjusted that widths of books to be kept is accounted for. One way to do this is to dedicate lower shelves for thicker books with more intermediary space, upper shelves for thinner books with lower intermediary space. There are IR emitters at every position, and an IR sensor mounted on the arm. These emitter-sensor pair from the required input for determination of position of the arm during movement and thus enable the arm to move to the desired location. After placing the book in the designated location, the arm then returns to the initial position and goes on standby. Similarly when a reader queues for a particular book, the address is searched in the system and the arm then goes to this address and retrieves the book for the user. This is a very simple extension of the above described process. A DC power supply (24 V, regulated and linear) is used to supply power to the microcontroller as well as several motors and drivers.

## 4 Experimental Results

The power supply was tested and was supplying a DC regulated power output (24V). The L298 motor drivers, connected to the arduino board were tested to run the motors. The voltage fluctuation observed was satisfactory, *viz.* 19-34 V. The operation of the IR sensors was also tested by connecting them to the relay circuit (Primary: 6V, 0.1A; Secondary: 240V 50Hz, 6A) and was successful. After this they were successfully interfaced with the Arduino. The testing of motor operation according to sensor input, which is the actual concept on which the project is based, was also done with satisfactory results. After assembly of the entire system, testing was carried out and the system could recognize books and retrieve them.

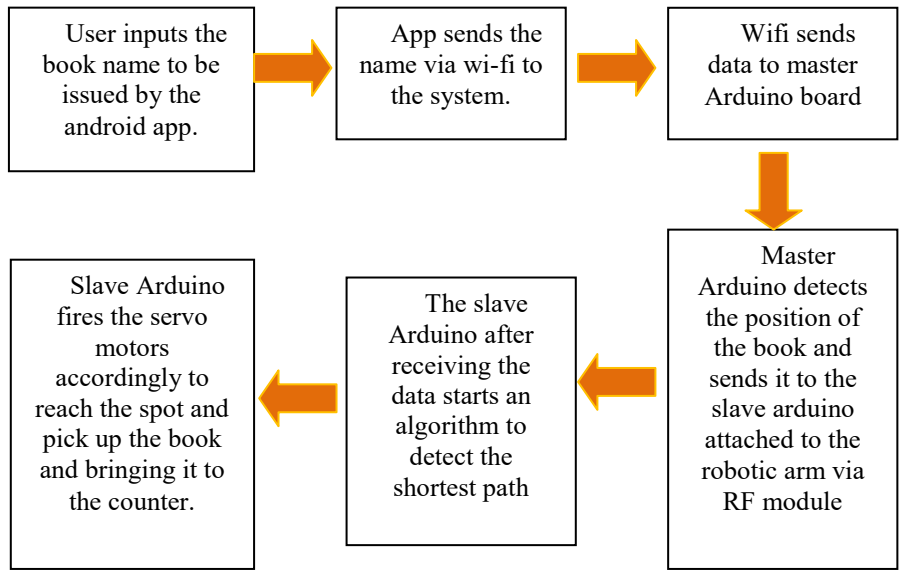

**Fig. 2.** Block diagram of the proposed work

## 5 Conclusion

The proposed system has brought an efficient way to manage libraries. One of the most notable things about it is its versatility, as it can easily be appended to any existing library, and the cost is proportional to the task load. Smaller libraries can be handled by cheaper microcontrollers whereas large scale management may be done using PLCs (especially if the system is applied to maintain store houses in industries). The system can be effectively implemented in medium and large sized libraries for simple and efficient management. It can also find application in other inventory management with minor modifications in programming. The described system reduces time, labor and brings about a safe and innovative way to manage libraries.

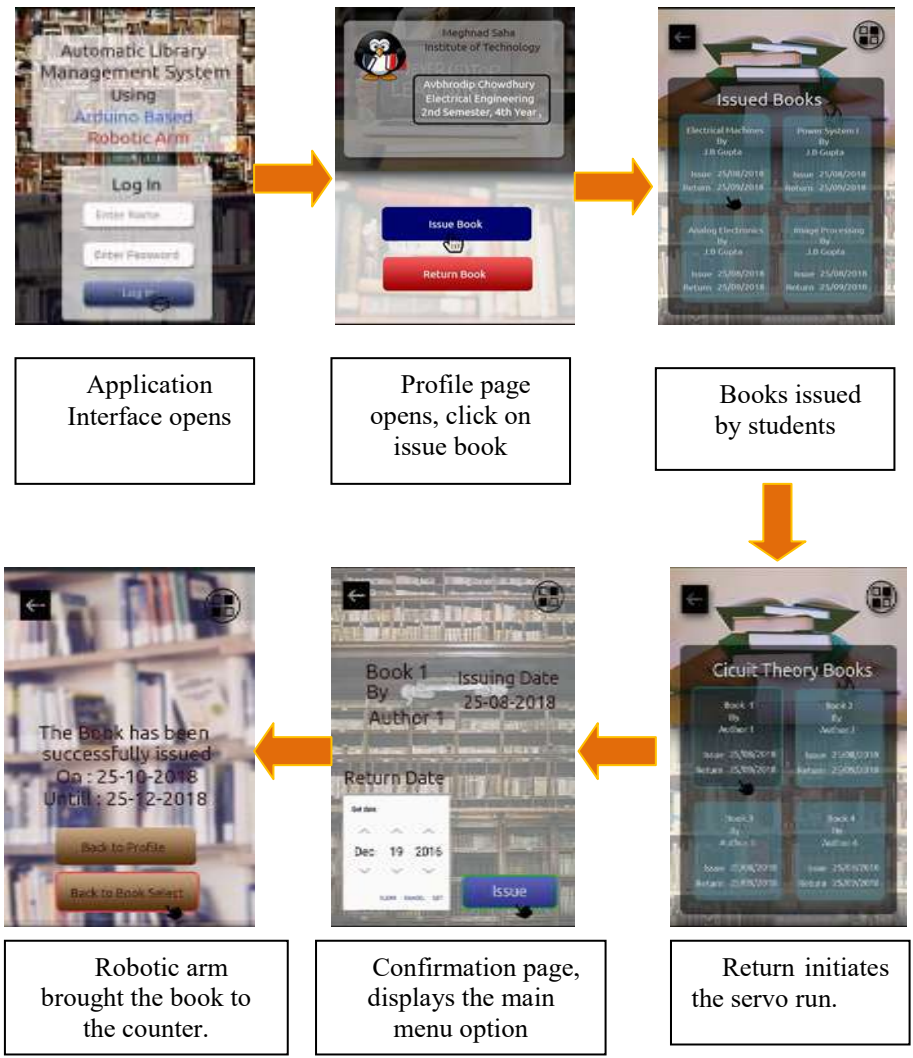

**Fig. 3.** Testing results of proposed system